\documentclass[runningheads]{llncs}
\usepackage[T1]{fontenc}
\usepackage{graphicx}
\usepackage{hyperref}
\usepackage{amsmath}
\usepackage{booktabs}
\usepackage{multirow}
\usepackage{xcolor}
\begin{document}
\title{Can MLLMs generate human-like feedback in grading multimodal short answers?}
\titlerunning{MMSAF}
%
\author{Pritam Sil\inst{1} 
\and
Pushpak Bhattacharyya\inst{1}
\and Pawan Goyal\inst{2}
\and Ganesh Ramakrishnan\inst{1}
}
%
%
%
\maketitle              
\begin{abstract}
In education, the traditional Automatic Short Answer Grading (ASAG) with feedback problem has focused primarily on evaluating text-only responses. However, real-world assessments often include multimodal responses containing both diagrams and text. To address this limitation, we introduce the Multimodal Short Answer Grading with Feedback (MMSAF) problem, which requires jointly evaluating textual and diagrammatic content while also providing explanatory feedback. Collecting data representative of such multimodal responses is challenging due to both scale and logistical constraints. To mitigate this, we develop an automated data generation framework that leverages LLM hallucinations to mimic common student errors, thereby constructing a dataset of 2,197 instances. We evaluate 4 Multimodal Large Language Models (MLLMs) across 3 STEM subjects, showing that MLLMs achieve accuracies of up to 62.5\% in predicting answer correctness (correct/partially correct/incorrect) and up to 80.36\% in assessing image relevance. This also includes a human evaluation with 9 annotators across 5 parameters, including a rubric-based approach. The rubrics also serve as a way to evaluate the feedback quality semantically rather than using overlap-based approaches. Our findings highlight which MLLMs are better suited for such tasks while also pointing out to drawbacks of the remaining MLLMs.

\keywords{MMSAF  \and MLLM \and Multimodal Answer Grading}
\end{abstract}
\section{Introduction}
\label{Introduction}
In educational assessments, the traditional Automatic Short Answer Grading (ASAG) with feedback problem focuses on developing scalable solutions for automatically grading short textual responses along with feedback. Such feedback is important~\cite{DEEVA2021104094}   for a student's growth. Moreover, corrective, motivational and informative feedback can drastically speed up a student's learning process and help the student develop an innate curiosity.

However, assessments often contain open-ended responses that have both a textual part and a diagram. These questions play a crucial role in assessments, as they require students to express their understanding through textual and visual elements, leading to deeper levels of cognitive engagement. While in some scenarios, students are required to draw such diagrams (E.g. ray diagrams in physics), in others, they are required to label certain aspects of it (E.g., a diagram of a body part, such as the heart in biology). In such a diagram, each small component, such as an arrow, holds a semantic value that contributes to evaluating the answer as a whole. Grading such responses at scale while providing individual feedback to students is often difficult, especially in classrooms with a low teacher-to-student ratio~\cite{Burrows2015TheEA}. 

This leads to the question: Can we develop scalable assessment tools that can assist teachers in evaluating multimodal responses while providing quality feedback? To facilitate research towards developing such systems, we define the Multimodal Short Answer grading with Feedback (MMSAF) problem. The problem is challenging for MLLMs as it tests their diagram understanding, natural language comprehension, and comparative  reasoning capabilities.

Our contributions are as follows -
\begin{enumerate}
    \item Introduction of the MMSAF problem, along with a dataset of 2,197 instances. (Section~\ref{Multimodal Short Answer Feedback (MMSAF) problem}) 
    \item MMSAF-DGF, a framework to generate an MMSAF dataset from a given set of standard label questions and reference answer pairs. It simulates common student errors using LLM hallucinations as a tool. (Section~\ref{Multimodal Short Answer Feedback (MMSAF) Dataset})
    \item A rubric-based approach to evaluate the quality of feedback coupled with extensive zero-shot evaluation on 4 existing MLLMs across 3 different STEM subjects.  We observe the following: (Section~\ref{Evaluation of LLM Generated Feedback})
    \begin{itemize}
        \item Accuracy of up to 62.5\% using existing MLLMs on generating the correctness level of a response (correct/partially correct/incorrect).
        \item  Accuracy of up to 80.36\% using existing MLLMs on generating the image relevance level (relevant/irrelevant) of a diagram in a response.
        \item Human evaluations indicate feedback generated by such MLLMs is closely related to how teachers provide feedback to students.
    \end{itemize}
\end{enumerate}
\section{Related Work}
In recent years, there has been growing interest from both the natural language processing (NLP) and education research communities in the task of Automatic Short Answer Grading (ASAG) with feedback. A notable milestone in this direction came in 2022, when Filighera et al.~\cite{filighera-etal-2022-answer} introduced the first dataset for ASAG with feedback problem. This bilingual dataset focused on short textual responses to questions across various topics, primarily in computer science. However, the dataset was limited to only about 2,000 responses, and lacked diversity across different  engineering disciplines.

To address these shortcomings, Aggarwal er al.~\cite{aggarwal2024iunderstandigot} introduced the EngSAF dataset, which contained about 5,800 student answers drawn from 25 different engineering courses spanning multiple subfields. This dataset laid the groundwork for more robust benchmarking and model development in the ASAG with the feedback task.

Following this, research began to shift towards more advanced methods for the ASAG with the feedback problem. While earlier approaches primarily leveraged carefully crafted prompts, Fateen at al.~\cite{fateen2024scoresmodularragbasedautomatic} proposed a retrieval-augmented generation (RAG) based approach to enhance response quality and contextual relevance.

Any feedback generated by the model serves as a way to explain its grading rationale, which can add a layer of explainability to the grading task. Li et al.~\cite{li-etal-2023-distilling} introduced the Automated Explainable Student Response Assessment (AERA) framework, which generates scoring rationales using ChatGPT. AERA demonstrated rationales comparable in quality to human explanations and achieved a Quadratic Weighted Kappa (QWK) score of 11\% on benchmark datasets. Similarly, Tornqvist et al.~\cite{tornqvist-etal-2023-exasag} presented ExASAG, an explainable grading framework that integrates SHAP (Shapley Additive exPlanations) with SciBERT to introduce interpretability to the process of grading such assignments.

While these efforts significantly advanced the field of text-based ASAG, they fall short in handling multimodal responses containing only visual models or diagrams. Visual models, as part of student answers, play an important role in demonstrating the student's proficiency level on a particular concept. To address this, Leong et al.~\cite{Cheeweeetal} and later Sagherian et al.~\cite{Sagherianetal} proposed automated grading systems for scientific visual models, evaluated on proprietary datasets from the Educational Testing Service (ETS). However, these visual models were created using predefined shapes such as boxes, arrows, and fish, limiting student expressiveness and creativity. Similarly, Lee et al.~\cite{lee2023nerifgpt4vautomaticscoring} showed that GPT-4V is capable of evaluating such visual models. However, none of these works deal with the scenario of evaluating responses containing both text and diagrams. 

To bridge this gap and handle real-world scenarios where students provide multimodal responses that include textual answers accompanied by diagrams, we introduce the MMSAF problem and explain why it is challenging for MLLMs to evaluate such responses in the next section.

\section{The Multimodal Short Answer Grading with Feedback (MMSAF) Problem}
\label{Multimodal Short Answer Feedback (MMSAF) problem}

\begin{figure*}
    \centering
    \includegraphics[scale=0.25]{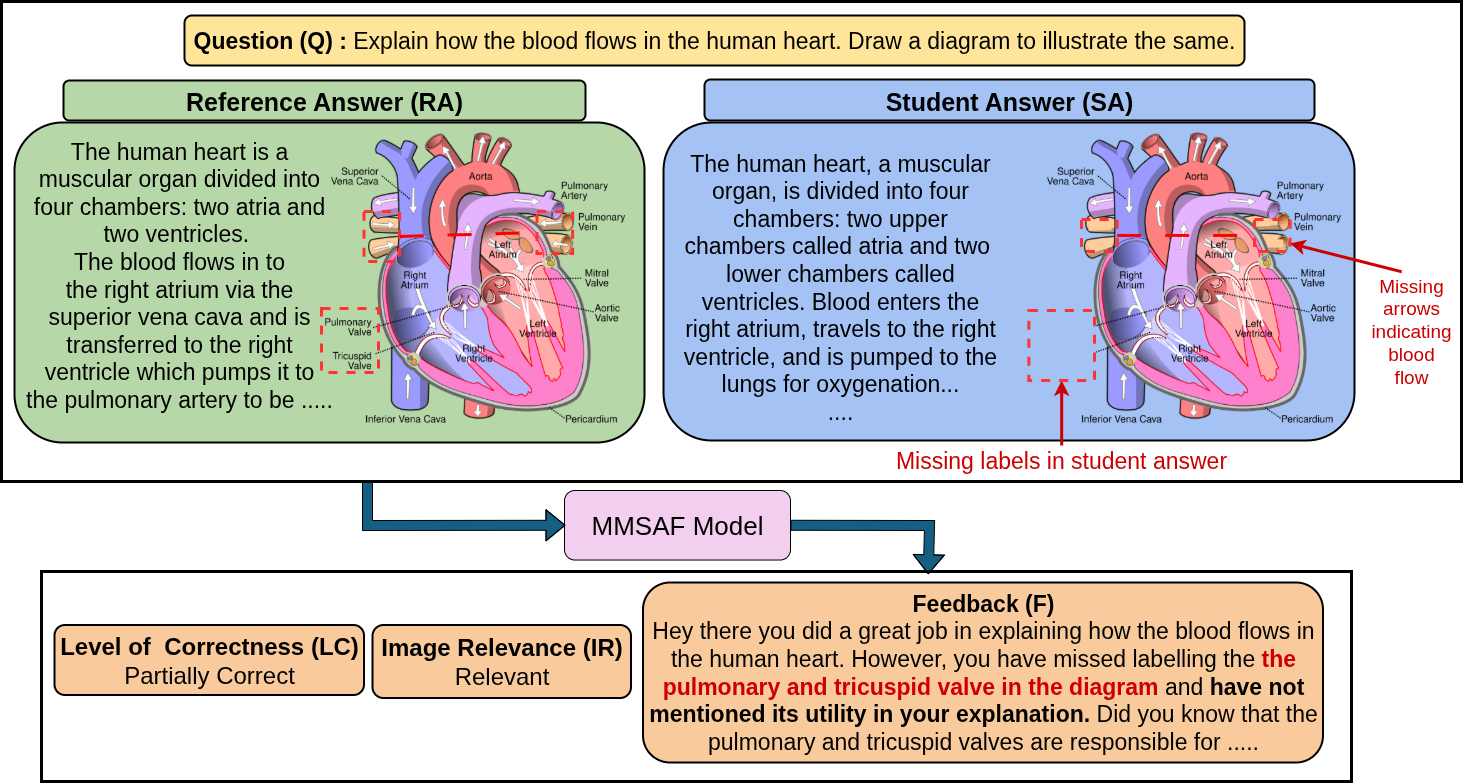}
    \caption{Illustration of the MMSAF problem with an example. (Heart diagram taken from \href{https://edurev.in/t/131714/STRUCTURE-OF-HUMAN-HEART}{edurev.in})}
    \label{MMSAF_Problem}
\end{figure*}

We introduce the Multimodal Short Answer Grading with Feedback (MMSAF) problem, which focuses on evaluating student responses containing both textual and visual content. Given a Question (Q), a Reference Answer (RA), and a Student Answer (SA), the objective is to assign a Level of Correctness (LC) label, an Image Relevance (IR) label, and a Feedback (F) which provides rationale behind the LC and IR labels.

Figure~\ref{MMSAF_Problem} provides an illustrative example. In the given question, the student is asked to \textit{explain} the flow of blood in the human heart \textit{using a diagram}. The question demands a supporting diagram, which adds clarity and depth to the textual response. Note that creating such diagrams using only simple visual models is difficult, making it significantly more challenging than standard visual model scoring tasks. This highlights the motivation for introducing this problem.
\subsection{Problem Formulation}
\label{Problem Formulation}
We decompose the MMSAF problem into two sub-problems: a classification task for LC and IR labels, and a feedback generation task.
\begin{itemize}
        \item \textbf{Classification Task: } Given the SA, the MLLM $M$ must evaluate it with respect to the given Q and RA and determine its correctness level that is correct (C) or partially correct (PC) or incorrect (I): 
        \[
        (Q,SA,RA)\xrightarrow{M}\{C,PC,I\}
        \]
    
    Similarly, given the SA, the model has to determine whether the image in the student response is relevant (R) or irrelevant (IRel) to the given Q and RA. This is to determine whether the diagram is adding value to the whole response or not. 
    \[(Q, SA, RA) \xrightarrow{M} \{\text{R, IRel}\}\]
    
    \item \textbf{Feedback Generation Task: } Here given the SA, the MLLM ($M$) has to evaluate it with respect to the Q and SA and generate a human like feedback. The feedback should point out the errors present in SA, ways to mitigate them and additional information which might assist the student. This acts as a small explanation as to why the LC and IR levels were assigned by the model to the SA, and increases trust in the scoring systems. Thus, it can be formulated as :
    \[(Q, SA, RA) \xrightarrow{M} \text{Feedback}\]

\end{itemize}

So, \textbf{why is this challenging for MLLMs?} The MMSAF problems require the model to understand the semantics present in both the text and diagram. This involves the model to perform the following -
\begin{itemize}
    \item \textbf{Natural language understanding:} The model has to analyse the semantics of the student response, including topic-specific details. This means the model must have prior knowledge of educational data to perform this task effectively \cite{peris-etal-2022-knowledge}.
    
    \item \textbf{Diagram understanding:} The model has to interpret and capture finer details present in the diagram, such as arrows, labels and even topic-specific objects, such as lenses in the case of physics. It depends on their ability to make use of their background knowledge as shortcuts to identify and reason about the relational information \cite{Giledereli2024DoVM}. 
    
    \item \textbf{Comparative reasoning:} Now that it has analysed the text and diagram present in both the SA and RA, it has to now compare between them. Thus, the model's comparative reasoning ~\cite{yu-etal-2023-pre} abilities play a crucial role in generating corrective, motivational, and informative feedback. 
    
\end{itemize}
In addition to these technical challenges, collecting data representative of such responses is often a significant challenge. This involves collaborations with different educational institutions, which may take a longer time to evolve, a lack of individualised feedback provided to students, and addressing privacy concerns related to sharing such data. To reduce the delay in curating such datasets, we propose an MMSAF-DGF in the next section. 

\section{MMSAF dataset generation framework (MMSAF-DGF)}
\label{MMSAF dataset generation framework}
MMSAF-DGF provides a  solution to generate the MMSAF dataset by leveraging standard questions and answer pairs, as well as LLM hallucinations. This reduces the time and monetary effort required in curating such datasets.

\subsection{MMSAF Dataset}
\label{Multimodal Short Answer Feedback (MMSAF) Dataset}
We formally define the \textbf{MMSAF dataset} where each data point is composed of 7 elements, namely: Q, RA, SA, LC, IR, F and Rubrics for error detection in feedback (FR). As per the problem, given a (Q, RA, SA) tuple, the task is to generate feedback that evaluates both the correctness and image relevance of the student's answer. Moreover, the rubric-based feedback evaluation strategy represented by the FR value 
serves the following purpose:

\begin{itemize}
    \item A way to semantically evaluate feedback quality rather than relying on existing metrics such as BLEU~\cite{10.3115/1073083.1073135} or ROUGE~\cite{Lin2004ROUGEAP}, which rely on word overlap.

    \item An approach towards reducing annotation cost by employing both humans and the LLM-as-a-judge paradigm to evaluate feedback quality.

\end{itemize}
Another aspect pertaining to FR is that it is specific to a particular datapoint and not shared, since the errors introduced in the responses are all question-specific. Note that in each datapoint, the Q and RA pairs are standardised, indicating that they have been carefully curated by subject matter experts (SMEs). The challenge is to generate the SA, LC, IR, F and FR components. Before we explain how the same is done, we explain how LLM-based hallucinations can be used to mimic student errors.

\subsection{Leveraging LLM hallucinations to simulate student errors}
\label{Leveraging LLM hallucinations to simulate student errors}
In this subsection, we elaborate the psychology behind common errors students make in examinations. Marsh et al.~\cite{Marsh_Eliseev_2019} primarily categorises them as follows:

\begin{itemize}
    \item \textit{Errors made with confidence:} This happens when students answer questions confidently but are unaware of the answers they provide. This means that the student is either confidently placing a correct fact as the response or making up answers to attempt it.
    E.g. “Haemoglobin consists of \underline{\textcolor{red}{calcium}} and globin.” Here, given the context, calcium is an incorrect response, and the student is unaware of it.
    \item \textit{Misunderstanding:} Such errors are often made by students when they are unable to understand the question properly or even the related concept. This indicates that the provided answer again contains some partially correct fact. E.g. ``Bats shout at objects, hence \textcolor{red}{mammals can too perform echolocation}'' Here, the student misunderstands the correct fact that bats have special organs to enable echolocation.

    \item \textit{Conceptual Change:} These kinds of errors are made when the student has understood a particular concept but is unable to apply it in a different context. This means that the student is confused and ends up using a fact or concept that is unrelated to the problem. E.g. ``Diffusion happens because the oxygen and carbon dioxide molecules \textcolor{red}{move willingly.}'' Here, diffusion occurs due to a difference in concentrations, but the student has written a wrong fact.
\end{itemize}
 We argue that LLM hallucinations, particularly factual fabrication and inconsistency, can be used as a tool to simulate such responses. As per Huang et al.~\cite{10.1145/3703155}, factual inconsistency involves using a correct fact in a wrong context, and factual fabrication involves making up incorrect facts. This phenomenon can be easily used to mimic student responses, as shown in the earlier examples. Using this notion, we explain how MMSAF-DGF generates an MMSAF dataset. 

\subsection{Overview of MMSAF-DGF}
\begin{figure*}
    \centering
    \includegraphics[scale=0.3]{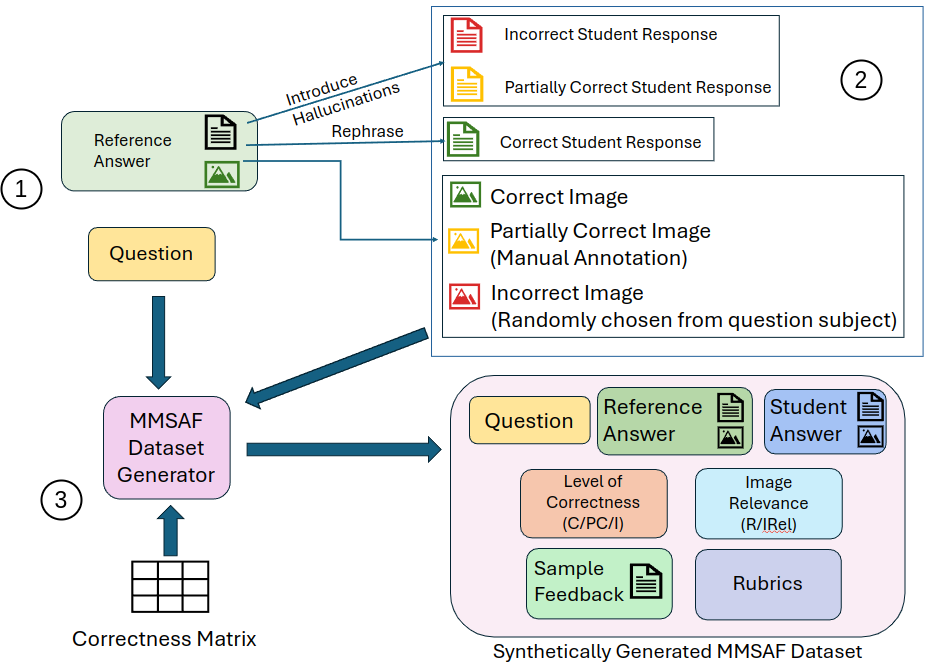}
    {\small
  \\C - Correct
  PC - Partially Correct
  I - Incorrect
  R - Relevant
  IRel - Irrelevant
  }
    \caption{Overview of the MMSAF dataset generation framework (MMSAF-DGF)}
    \label{DataGenerationFramework}
\end{figure*}
Figure~\ref{DataGenerationFramework} illustrates the overall workflow of MMSAF-DGF. The framework consists of the following steps:

\textbf{Step 1:} It takes in a set of standard questions and reference answer pairs. Note that the reference answer contains both a text and a figure or diagram.

\textbf{Step 2:} The framework introduces errors to the correct answers to generate the partially correct and incorrect student responses. We take a divide-and-conquer approach by introducing errors in the text and visual part separately and then combining them using a correctness matrix to generate a data point.

For the textual part, hallucinations are introduced to generate the partially correct and incorrect responses, while the correct answers are rephrased to generate the correct student response. The justification for this methodology has already been motivated in Section~\ref{Leveraging LLM hallucinations to simulate student errors}. Gemini~\cite{geminiteam2024geminifamilyhighlycapable} was employed to generate hallucinations due to its free availability. To assess its effectiveness in simulating student errors, we evaluated 50 instances each for Partially Correct (PC) and Incorrect (I) responses. Our results show that Gemini successfully produced hallucinated responses in 86\% of the Incorrect cases and 80\% of the Partially Correct cases. 

For the provided diagram, we treat it as a correct response diagram. Partially correct images are generated by removing labels and objects, replacing closely related objects or swapping labels to generate a , and randomly assigning another image from the subject pool to generate the incorrect image. While LLMs can be used to specify what needs to be modified in an image, actually modifying the image accordingly remains challenging, especially when using openly available models. Hence, we take a manual approach for simulating such errors as an initial step. Additionally, we focus only on simple mistakes made by students, such as mislabelling or omitting labelling certain objects, incorrect use of directional entities (such as arrows in free-body diagrams), and even incorrect use of technical objects (such as lenses).

\begin{table}[h!tbp]
  \centering
  \begin{tabular}{c|c|c}
    \hline \hline
    \textbf{SA Text} & \textbf{SA Image} &
    \textbf{Overall Correctness Label} \\
    \hline \hline
    C   & C       & C   \\
    C   & PC/I    & PC  \\
    PC  & C/PC    & PC  \\
    PC  & I       & I   \\
    I   & C       & PC  \\
    I   & PC/I    & I   \\
    \hline \hline
  \end{tabular}
  \caption{Matrix for determining the Level of Correctness}
  \label{correct_m}
\end{table}

\textbf{Step 3:} The C, PC and I textual and visual components are combined to generate the final student response using the provided correctness matrix (Table~\ref{correct_m}). When introducing hallucinations, we record the incorrect fact introduced and formulate it as a question for the rubrics and as a text for the sample feedback. 

Using this framework, we generate a dataset of \textit{2,197 instances} using \textit{181 high school-level questions across physics, chemistry, and biology}.  In the next section, we first justify the choices of the 4 MLLMs in the next section and use the generated dataset to determine whether current MLLMs can provide human-like feedback for the MMSAF problem.

\section{Evaluation of LLM Generated Feedback}
\label{Evaluation of LLM Generated Feedback}
The goal of this evaluation is to benchmark the zero-shot performance of existing MLLMs on this dataset and grade their capabilities. We first motivate our selection of existing, followed by their evaluation. To do so, 221 data points (130 from biology, 56 from chemistry and 35 from physics) were sampled randomly. These were then analysed by MLLMs, and their generated level of correctness, image relevance labels and feedback were collected and analysed in the later sections. The corresponding LC , IR and feedback values generated were collected and then analysed in the following subsections.

\subsection{LLMs in Consideration}
\label{LLMs in Consideration}
The MMSAF problem involves evaluating multiple images and text as a whole. As mentioned in Section~\ref{Problem Formulation}, this is a complex task that relies on the MLLM's diagram understanding, natural language understanding, and comparative reasoning capabilities. This indicates that only models that support multiple images as inputs, are well-versed in complex tasks, and have educational data in their prior knowledge can be used. Given this motivation, we select 4 such models: ChatGPT, Gemini, Pixtral, and Molmo. 

We consider ChatGPT, which has demonstrated strong performance in grading and educational tasks, including diagram-based scoring~\cite{lee2023nerifgpt4vautomaticscoring}, making it a suitable candidate for MMSAF. Similarly, in recent times, Gemini has also shown extensive performance in educational tasks such as answer grading and question generation\footnote{\url{https://edu.google.com/intl/ALL_in/ai/gemini-for-education/} (Last Accessed: 24-01-2026)}. In terms of open source models, we consider Pixtral since  it is trained to interpret diagrams while providing detailed and structured interpretations. While we consider Molmo, as it has been explicitly trained on educational data and has outperformed leading models, such as ChatGPT and Gemini, on 11 different academic benchmarks and in human evaluations\footnote{\url{https://allenai.org/blog/molmo }(Last accessed : 24-01-2026)}. 

\subsection{Analysis of Correctness and Relevance levels}
\label{Analysis of Correctness and Relevance levels}
To evaluate the performance of the models in predicting Level of Correctness and Image Relevance labels, we report macro-averaged accuracy values across each subject, namely physics, chemistry, and biology.
\begin{table*}[ht]
\centering
\begin{tabular}{l|ccc|ccc}
\toprule
\multirow{2}{*}{\textbf{Model}} & 
\multicolumn{3}{c|}{\textbf{Level of Correctness}} & \multicolumn{3}{c}{\textbf{Image Relevance}} \\
\cmidrule(lr){2-4} \cmidrule(lr){5-7}
& Physics & Chemistry & Biology & Physics & Chemistry & Biology \\
\midrule
ChatGPT & 48.57 &58.93 &47.29 &\textbf{77.14} &\textbf{80.36}  &\textbf{72.09}\\
Molmo &40.00 &39.29 &43.85 &34.29 &26.79 &29.23 \\
Pixtral & 51.43 &58.93 &49.23 &74.29 &60.71 &66.15 \\
Gemini  & \textbf{57.14} &\textbf{62.50} &\textbf{50.77} &\textbf{77.14} &51.79 &55.38 \\
\bottomrule
\end{tabular}
\caption{Accuracy of models on generating LC and IR levels}
\label{LC_IR_tab}
\end{table*}

As shown in Table~\ref{LC_IR_tab}, Gemini outperforms all other models across the evaluated metrics, suggesting higher reliability in predicting LC labels and fewer false positives. This is due to equal strength in the 3 capabilities (as in Section~\ref{Problem Formulation}) necessary to solve this task. When it comes to ChatGPT, it had a tendency to label most answers as “Partially Correct” class, which impacted its overall performance. Molmo, in contrast, exhibited a strong bias towards labelling answers as “Incorrect” indicating that it is not so adept at this complex task. Pixtral, although more lenient, frequently confused “Incorrect” and “Partially Correct” responses, resulting in reduced performance but close to that of Gemini. However, this behaviour indicates potential for performance improvement through fine-tuning. 

Table~\ref{LC_IR_tab} also presents the results for IR prediction. ChatGPT achieves the highest performance across all metrics, indicating strong reasoning and diagram understanding capabilities. However, Molmo frequently predicted most images as relevant, resulting in poor performance, indicating scope for improvement in its performance. Pixtral also suffered from false positives, while Gemini often misclassified relevant images as irrelevant.
Next, we analyze whether they feedback generated by them are close to human judgement or not.

\subsection{Evaluation Task for Experts}
The MLLM-generated feedback, comprising an average of 1758 words, was recorded and presented to 9 Subject Matter Experts (SMEs), comprising 3 experts each for the domains of physics, chemistry, and biology. 

\label{Evaluation Task for Experts}
The SMEs were instructed to score each feedback on a scale of 1 to 5 based on the following parameters - 
\begin{enumerate}

\item \emph{Fluency and Grammatical Correctness (FGE):} This metric denotes the fluency and grammatical correctness of the LLM-generated feedback. The idea is to check if the LLM-generated sentences are grammatically correct or not. A score of 1 denotes that the FGE level of the feedback is extremely poor, while a score of 5 indicates that it is excellent.

\item \emph{Emotional Impact (EI):} This metric is to check whether the LLM-generated feedback will have a positive impact on the student or not, that is, whether the feedback is more encouraging and assistive for the student or not. A score of 1 denotes a negative impact, while a score of 5 denotes a positive impact. 

\item \emph{Level of Feedback Correctness (LFC):} This metric is to determine whether the feedback has properly captured all the errors present in the student's answer. A score of 1 denotes that no error has been captured in the feedback, while a score of 5 denotes that all the errors have been captured in the feedback.

\item \emph{Error Mitigation in Feedback (EM):} This metric evaluates whether the feedback has properly addressed each and every error present in the student's answer and suggested ways to correct them. A score of 1 denotes no such error mitigation has been done, while a score of 5 denotes that all the ways necessary to correct all errors are present.

\item \emph{Rubrics for error detection (FR):} While traditional NLP metrics like ROUGE-2~\cite{ganesan2015rouge}, SCAReBLEU~\cite{post-2018-call}, METEOR~\cite{banerjee-lavie-2005-meteor}, and BERTScore~\cite{Zhang-2020-ICLR} rely on n-gram overlap, they often miss the semantic accuracy of feedback, particularly in identifying and addressing student errors. To address this, we propose a rubric-based evaluation that checks whether the feedback captures all relevant errors. Annotators assess each rubric as a True/False question based on the LLM-generated feedback.
 
\end{enumerate}
\subsection{Analysis of Expert Evaluation}
\label{Analysis of Expert Evaluation}
\begin{table*}[ht]
\centering
\setlength{\tabcolsep}{6pt}
\begin{tabular}{llccccc}
\toprule
\textbf{Subject} & \textbf{Model} & \textbf{FGE} & \textbf{EI} & \textbf{LFC} & \textbf{EM} & \textbf{FR} \\
\midrule
\multirow{4}{*}{Physics}
 & ChatGPT & \textbf{5.00} & \textbf{4.59} & \textbf{4.53} & \textbf{4.56} & \textbf{0.78} \\
 & Molmo   & 4.75 & 2.66 & 2.69 & 2.25 & 0.53 \\
 & Gemini  & 5.00 & 4.56 & 4.12 & 4.09 & 0.68 \\
 & Pixtral & 5.00 & 4.15 & 4.24 & 4.24 & 0.55 \\
\midrule
\multirow{4}{*}{Chemistry}
 & ChatGPT & \textbf{4.98} & \textbf{4.98} & \textbf{4.42} & \textbf{4.51} & 0.63 \\
 & Molmo   & 4.70 & 3.25 & 2.93 & 2.98 & 0.48 \\
 & Gemini  & 4.95 & 4.67 & 4.37 & 4.42 & \textbf{0.65} \\
 & Pixtral & 4.95 & 4.71 & 4.10 & 4.05 & 0.49 \\
\midrule
\multirow{4}{*}{Biology}
 & ChatGPT & 5.00 & 3.07 & 3.24 & 3.15 & 0.50 \\
 & Molmo   & 4.92 & 3.11 & 3.04 & 2.69 & 0.54 \\
 & Gemini  & 5.00 & 3.59 & \textbf{3.43} & 3.07 & 0.53 \\
 & Pixtral & \textbf{5.00} & \textbf{3.87} & 3.40 & \textbf{3.48} & \textbf{0.58} \\
\bottomrule
\end{tabular}
\caption{Average expert evaluation scores across subjects and models}
\label{Annot_All}
\end{table*}
Once the three SMEs had completed their respective evaluation tasks, all the scores for each metric were collected and averaged out to present the final data. 
Table~\ref{Annot_All} summarises the average ratings assigned by annotators to the feedback generated by different LLMs over various criteria mentioned in Section~\ref{Evaluation Task for Experts} for each and every subject. Inter annotator agreement values ranged from tentative to strong, except for LC and EM in Molmo's Physics and Chemistry outputs, where a senior SME (10+ years experience) consistently rated more strictly than junior annotators (<5 years), leading to lower agreement. However, all Physics and Chemistry SMEs agreed Molmo’s feedback was inadequate, due to hallucinations, harsh tone, poor subject knowledge, and self-contradictory statements. These qualitative observations are consistent with the scores as in Table~\ref{Annot_All}.  

\textit{Physics:} Physics questions involve interpreting abstract diagrams containing objects such as arrows, circles and rectangles. It also involves interpreting certain domain-specific objects such as lenses and mirrors. Apart from diagrams, such questions involve calculations and physics concepts based on reasoning. ChatGPT outperformed others in all areas, with SMEs highlighting its strength in identifying calculation errors and providing step-by-step explanations. The scores from Table~\ref{Annot_All} indicate that it is most closely aligned to how humans provide feedback in Physics. However, in contrast, Molmo often hallucinated, mislabeled correct answers, and struggled with reasoning, though it handled concrete diagrams reasonably well. Pixtral delivered structured feedback with clearly structured explanations, aligning with its training data, where it had to interpret graphs and diagrams and provide structured analysis.

\textit{Chemistry:} Chemistry questions involve domain-specific chemical formulas as part of their questions and diagrams. They also include interpreting graphs. Some diagrams involve interpreting domain-specific objects such as a beaker, a scientific fork and others. When compared to other models, ChatGPT performs the best on all parameters, which indicates that it also provides human-like feedback in chemistry. Additionally, SMEs note that ChatGPT provides detailed explanations. The reason is the same as for physics questions, as these questions test the reasoning skills of such MLLMs and how well they can interpret abstract diagrams. However, they also point out that Molmo's tone was too direct for students and did not provide proper explanations for its assigned labels, particularly in terms of correctness and image relevance. This is because Molmo is not particularly trained to interpret such domain-specific notations and symbols involved in chemistry.

\textit{Biology:} Biology questions are more factual, involving real-life diagrams such as a heart. Interpreting such diagrams is much more complex as it requires interpreting domain-specific objects coupled with minute details such as arrows and labels which add semantic value to the diagram. MLLMs trained on such data will perform better in identifying key concepts. Pixtral excelled, providing structured, concept-based feedback and maintaining a polite tone, leading to higher emotional impact scores. This is also attributed to its training data, which consists of chart interpretation, allowing it to capture minute semantic details present in the diagrams. Although Molmo demonstrated a decent understanding of the diagram, it failed to connect errors to relevant concepts, highlighting a gap in its reasoning. ChatGPT and Gemini performed well, but not as effectively as Pixtral, which is attributed to its inability to interpret minute details present in such diagrams.

To summarise, ChatGPT is most aligned towards human judgement in generating feedback for physics and chemistry due to its strong reasoning skills and diagram interpretation skills. However, Pixtral is more effective in biology, which is attributed to its ability to capture minute elements present in such biological diagrams. It is also due to the fact that it provides detailed, empathetic feedback and structured analysis. Again, SMEs noted Molmo’s feedback often included Chinese alphabets, was overly direct, and sometimes mislabeled correct answers, suggesting a need for fine-tuning. 

\section{Conclusion}
Grading of multimodal responses and providing individualised feedback at scale is challenging. To aid in designing scalable solutions to this problem,
this work introduces the MMSAF problem, coupled with a data generation framework (MMSAF-DGF) that can generate such datasets using standard Q and RA pairs.  The framework leverages LLM hallucinations as tools to simulate erroneous student responses. While such synthetic datasets always enable a starting point, collection and evaluation on real-life data ensures reliable deployment of such models in specific use cases. Notetheless, we use this framework to generate a dataset of 2,197 instances, built on questions from physics, chemistry, and biology textbooks used in high school. 

We also establish a baseline using 4 MLLMs across all 3 subjects. Human evaluations over the generated feedback across 5 parameters indicate  that certain MLLMs in each subject are more aligned towards human judgment, indicating their potential usage in AI-assisted grading systems.  
%
%
%
\bibliographystyle{splncs04}
\bibliography{citations}
%




\end{document}